\documentclass[10pt,twocolumn,letterpaper]{article}

\usepackage[pagenumbers]{cvpr} 

\definecolor{cvprblue}{rgb}{0.21,0.49,0.74}
\usepackage[pagebackref,breaklinks,colorlinks,allcolors=cvprblue]{hyperref}
\usepackage{lipsum}
\usepackage{makecell} 
\usepackage{multirow}
\usepackage{xcolor}
\usepackage[table]{xcolor}
\usepackage{arydshln}

\usepackage{pifont}
\newcommand{\cmark}{\ding{51}} 

\definecolor{roborefgreen}{rgb}{0.92, 1.0, 0.92}
\definecolor{roborefgray}{gray}{0.95}
\definecolor{roborefblue}{rgb}{0.88,0.98,1}
\definecolor{roborefred}{rgb}{1, 0.9, 0.9}

\newcommand{\sota}[1]{\textbf{\textcolor{red}{#1}}}

\usepackage{xspace}
\makeatletter
\DeclareRobustCommand\onedot{\futurelet\@let@token\@onedot}
\def\@onedot{\ifx\@let@token.\else.\null\fi\xspace}
\def\eg{\emph{e.g}\onedot} 
\def\ie{\emph{i.e}\onedot} 
\def\cf{\emph{cf}\onedot} 
 \def\vs{\emph{vs}\onedot}

\makeatother

\def\confName{CVPR}
\def\confYear{2026}

\title{Seeing through Imagination: Learning Scene Geometry via\\ Implicit Spatial World Modeling}
\author{Meng Cao\textsuperscript{1}\footnotemark[1],~~Haokun Lin\textsuperscript{1}\footnotemark[1],~~Haoyuan Li\textsuperscript{2}\footnotemark[1],~~Haoran Tang\textsuperscript{3}\footnotemark[1],~~Rongtao Xu\textsuperscript{1,4},\\Dong An\textsuperscript{1},~~Xue Liu\textsuperscript{1},~~Ian Reid\textsuperscript{1}, ~~Xiaodan Liang\textsuperscript{1,2}$^{\dagger}$ \\
\textsuperscript{1}MBZUAI~~\textsuperscript{2}SYSU~~\textsuperscript{3}PKU~~\textsuperscript{4}Spatialtemporal AI\\
	{\small{\textsuperscript{*}Authors contributed equally to this research.~~\textsuperscript{\dag}Corresponding author.}}\\
	\textbf{\url{https://github.com/mengcaopku/MILO}}\\
}

\begin{document}

\twocolumn[{
	\renewcommand\twocolumn[1][]{#1}
	\maketitle
    \begin{center}
		\includegraphics[width=0.98\linewidth]{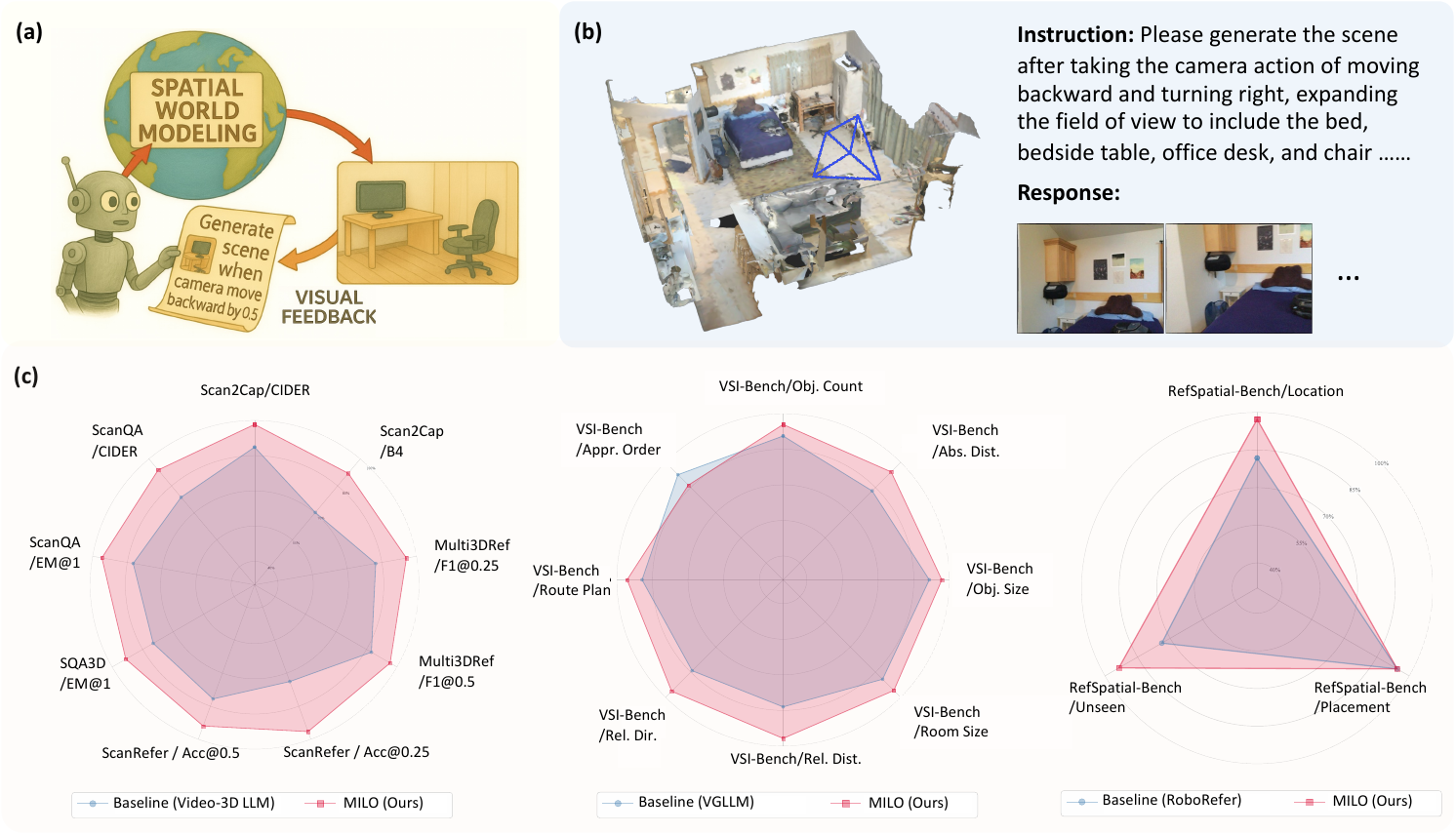}
		\captionsetup{type=figure}
     \caption{\textbf{(a) Conceptual illustrations} of our implicit spatial world modeling. \textbf{(b) Example case of GeoGen dataset}. Note that the point cloud scene is for illustrative purposes only, whereas raw video data is actually employed as input. \textbf{(c) Performance boost} of our MILO over different baselines including Video-3D LLM \cite{zheng2025video}, VGLLM \cite{zheng2025learning}, and RoboRefer \cite{zhou2025roborefer}.}
		\label{fig:teaser}
	\end{center}
}]
\begin{abstract}
Spatial reasoning, the ability to understand and interpret the 3D structure of the world, is a critical yet underdeveloped capability in Multimodal Large Language Models (MLLMs). Current methods predominantly rely on \emph{verbal} descriptive tuning, which suffers from visual illiteracy, \ie, they learn spatial concepts through textual symbols alone, devoid of connection to their visual manifestations. To bridge this gap, this paper introduces \textbf{MILO}, an \textbf{I}mplicit spat\textbf{I}aL w\textbf{O}rld modeling paradigm that simulates human-like spatial imagination. MILO integrates a visual generator to provide geometry-aware feedback, thereby implicitly grounding the MLLM's symbolic reasoning in perceptual experience. Complementing this paradigm, we propose \textbf{RePE} (\textbf{Re}lative \textbf{P}ositional \textbf{E}ncoding), a novel encoding scheme that captures relative camera-pose transformations, offering superior performance over absolute coordinate systems. To support the training, we construct GeoGen, a large-scale Geometry-aware Generative dataset with approximately 2,241 videos and 67,827 observation–action–outcome triplets. Experiments demonstrate that our approach significantly enhances spatial reasoning capabilities across multiple baselines and benchmarks, offering a more holistic understanding of 3D space. 
\end{abstract}
\section{Introduction} \label{sec:intro}
Spatial reasoning \cite{yang2025thinking,zhang2025flatland,deng20253d} is the cognitive process of understanding and interpreting the 3D structure of the physical world, \eg, computing metric quantities and inferring spatial relationships among entities. It lies at the core of numerous vision and robotics tasks such as autonomous driving \cite{shao2024lmdrive,sima2024drivelm}, embodied navigation \cite{zheng2024towards,lin2025navcot}, and robotic manipulation \cite{gao2024physically,nasiriany2024pivot,qu2025spatialvla}. Early attempts \cite{fu2024scene,zhang2024chatscene,hong20233d} commonly utilize 3D point clouds as input and align them with the representation space of Multimodal Large Language Models (MLLMs). Nevertheless, due to the limited ability of current MLLMs to directly process 3D data and the high cost of acquiring high-quality 3D assets, recent efforts have shifted toward learning 3D priors from multi-view images or videos \cite{zheng2025video,qi2025gpt4scene,zheng2025learning,fan2025vlm,chen2025sd}.

\begin{figure}[t]
	\centering
        \includegraphics[width=0.48\textwidth]{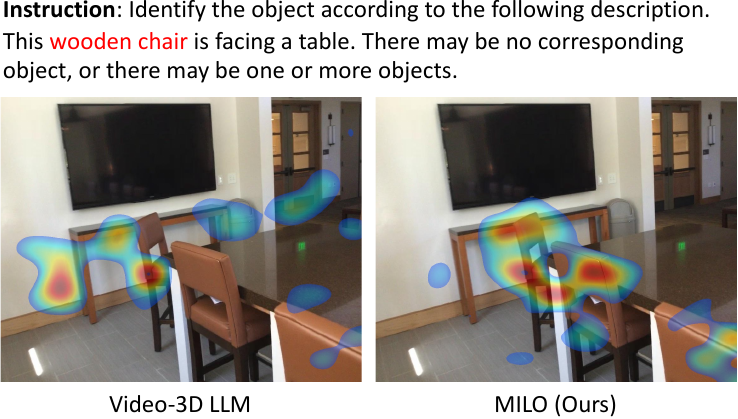}
     \caption{\textbf{Visualization of the attention scores} of the last generated token with respect to all visual tokens. The baseline MLLM \cite{zheng2025video} demonstrates visual illiteracy, \ie, failing to attend to the correct region of interest when generating responses. Refer to Table \ref{tab:gasRes} for detailed quantitative analysis and supplementary material for more visualization cases.}
	\label{fig:attentionCompare}
\end{figure}

To enhance the spatial reasoning capabilities, most existing methods follow a \emph{verbal descriptive tuning} paradigm \cite{liao2025improved,chen2025sd,zhang2025flatland,zheng2025learning,fan2025vlm,huang2025mllms}, \ie, they curate spatially-oriented datasets and instruct MLLMs to describe spatial properties (\eg, relative direction, object distance) in a single textual modality through supervised fine-tuning or reinforcement learning. Despite the progress, this descriptive tuning paradigm suffers from \textbf{visual illiteracy}, \ie, it relies solely on the textual symbolic supervision and is never exposed to how spatial transformations actually manifest in the visual domain.  Consequently, MLLMs often fail to attend to the correct region of interest. In Figure \ref{fig:attentionCompare}, we visualize the attention scores of the last generated token with respect to all visual tokens. The baseline model, Video-3D LLM \cite{zheng2025video}, fails to properly attend to the region containing the target ``\texttt{wooden} \texttt{chair}", indicating that the current training scheme lacks cross-modal grounding between spatial semantics and visual perception. In contrast, for human spatial cognition, imagining and mentally simulating spatial structures is an intuitive process that grounds reasoning in perceptual experience rather than symbolic abstraction. 

Therefore, we propose an i\textbf{M}plicit spat\textbf{I}a\textbf{L} w\textbf{O}rld modeling (\textbf{MILO}) paradigm that complements current verbal descriptive tuning with visual \emph{generative} tuning. As shown in Figure \ref{fig:teaser}(a), a \emph{generator} is integrated after MLLMs to yield visual feedback supervision under geometry-aware transformation instructions, such as viewpoint changes in Figure \ref{fig:teaser}(b). In this way, we implicitly bridge symbolic spatial reasoning with perceptual grounding, enabling MLLMs to internalize how geometric transformations manifest in the visual domain. Notably, a recent work Ross3D \cite{wang2025ross3d} also adopts a form of visual generative tuning; however, it is primarily confined to the tasks of masked visual prediction and BEV reconstruction. Moreover, it lacks explicit transformation instructions fed into the MLLMs, which prevents the models from perceiving and reasoning about the underlying geometric transformations they perform.

%

To inject 3D awareness into pre-trained MLLMs, recent works \cite{zheng2025video,zhu2024llava} introduce 3D positional embeddings that encode patch-wise 3D coordinates in the world coordinate system and bind them to 2D semantic patch tokens \cite{radford2021learning} to form ``3D patches." However, such \emph{absolute} positional encodings inherently rely on a global \emph{video-specific} coordinate system, making them sensitive to arbitrary world-coordinate choices and thus hindering generalization across datasets or camera setups. To this end, we propose a \textbf{Re}lative \textbf{P}ositional \textbf{E}ncoding (\textbf{RePE}) scheme that captures \emph{relative} camera-pose variations between adjacent frames based on intrinsic and extrinsic parameters. Unlike absolute encodings, our RePE is coordinate-system agnostic and explicitly models the geometric relationships among multiple views (\eg, how cameras are positioned and oriented with respect to one another).

To facilitate the training of our MILO, we construct a Geometry-aware Generative (GeoGen) dataset, which models interaction outcomes under various geometric transformations (\ie, novel view synthesis and trajectory generation). As shown in Figure \ref{fig:teaser} (b), GeoGen consists of input visual frames (\ie, observations), the executed instructions (\ie, actions), and the resulting frames (\ie, outcomes). To ensure extensive coverage, we construct the GeoGen dataset with videos sourced from both scanned 3D scenes and the web, comprising 2,241 videos and 267,827 observation–action–observation triplets. By conducting our visual generative post-training on the GeoGen dataset followed by verbal descriptive tuning, we validate the consistent effectiveness of MILO across multiple baseline models, including Video-3D LLM \cite{zheng2025video}, VGLLM \cite{zheng2025learning}, and RoboRefer \cite{zhou2025roborefer}. As shown in Figure \ref{fig:teaser} (c), when built upon Video-3D LLM, our MILO achieves a 3.2\% absolute gain on the Acc$@$0.25 metric of ScanRef \cite{chen2020scanrefer} dataset.

In summary, our contributions are in three-folds:
\begin{itemize}[topsep=0pt, partopsep=0pt, leftmargin=13pt, parsep=0pt, itemsep=3pt]
    \item We introduce MILO, a novel implicit spatial world modeling paradigm that complements existing verbal descriptive tuning by integrating a visual generative tuning to provide geometry-aware feedback.

    \item RePE is proposed to explicitly model relative camera-pose variations for coordinate-agnostic positional encoding, instead of relying on absolute 3D coordinates.

    \item We construct GeoGen dataset for geometry-aware generative post-training, which yields consistent performance gains across multiple baselines and benchmarks.
\end{itemize}

\section{Related Work} \label{sec:related}
\subsection{Spatial Reasoning}

Despite substantial progress in general scene understanding, MLLMs still struggle with 3D spatial reasoning \cite{xu2025defining,zha2025enable,cai2025has,gholami2025spatial,yin2025spatial,ray2024sat,zhang2025open3dvqa,ma20243dsrbench,zhang2024sphere,lee2025perspective}, which typically requires geometric and relationship reasoning in three-dimensional space. To mitigate this, preliminary approaches rely on the input point clouds \cite{fu2024scene,hong20233d,zhang2024chatscene} or depth images \cite{liu2025ssr,cai2024spatialbot,daxberger2025mm,chen2025sd,qi2025gpt4scene}. However, acquiring high-quality 3D assets is costly, and lifting 2D data into 3D \cite{miao2025towards,chen2024spatialvlm} forms often involves complex pipelines, which severely impedes scalability and generalization. Therefore, recent studies have attempted to understand the 3D world directly from video data by curating large-scale, spatially oriented instruction datasets. To inject 3D priors, VGLLM \cite{zheng2025learning} and VLM3R \cite{fan2025vlm} augment the input with geometric embeddings extracted from pre-trained 3D foundation models \cite{wang2025vggt,wang2025continuous}. Another line of work, including 3DRS \cite{huang2025mllms} and ThinkWith3D \cite{chen2025think}, adopts a distillation strategy that leverages VGGT \cite{wang2025vggt} as a teacher model to calibrate the visual embeddings of MLLMs. However, existing spatial instruction tuning remains confined to the verbal level, and how the visual hidden states of MLLMs contribute to the final response remains unclear. Therefore, we propose a visual generative tuning paradigm that supervises MLLMs with visual feedback, \ie, the visual states resulting from transformation instructions, which enables MLLMs to establish a human-like implicit spatial world modeling.

\subsection{World Models} 

World Models \cite{sutton1991dyna,ha2018recurrent,hafner2019dream,hafner2020mastering} are initially proposed to learn compact latent representations for visual dynamics prediction. Recently, with the remarkable generative capabilities of video diffusion models, emerging studies have begun exploring their potential as interactive world models capable of simulating complex and controllable environments \cite{alonso2024diffusion,bruce2024genie,che2024gamegen,valevski2024diffusion,xiao2025worldmem}. By conditioning on the agent's actions, these models can synthesize responsive and dynamic virtual worlds. On the other hand, such agent-environment interaction can also serve as an auxiliary prediction task for the agent's policy \cite{zhang2025agent,yu2025dyna}. Early Experience \cite{zhang2025agent} leverages future states to enhance the agent by constructing internal representations of environmental dynamics, while Dyna-Think \cite{yu2025dyna} integrates world model simulation directly into the agent's reasoning process. For spatial reasoning tasks, WorldLM \cite{zhang2025can} employs world model \cite{blattmann2023stable} as dynamic state encoders to capture temporal variations. Although effective, it suffers from limited interpretability and incurs additional input embeddings. In contrast, our MILO framework actively enables geometry-aware interactions between MLLMs and the environment, thereby learning from the visual feedback of such interactions. The recent study Ross3D \cite{wang2025ross3d} also adopts a generative fine-tuning paradigm, but it lacks explicit environment interaction since no concrete transformation instructions are provided to MLLMs. Moreover, it is confined to fixed masked visual prediction and BEV reconstruction tasks, resembling more a form of self-supervised and supervised representation learning.
\begin{figure*}[t]
	\centering
        \includegraphics[width=0.99\textwidth]{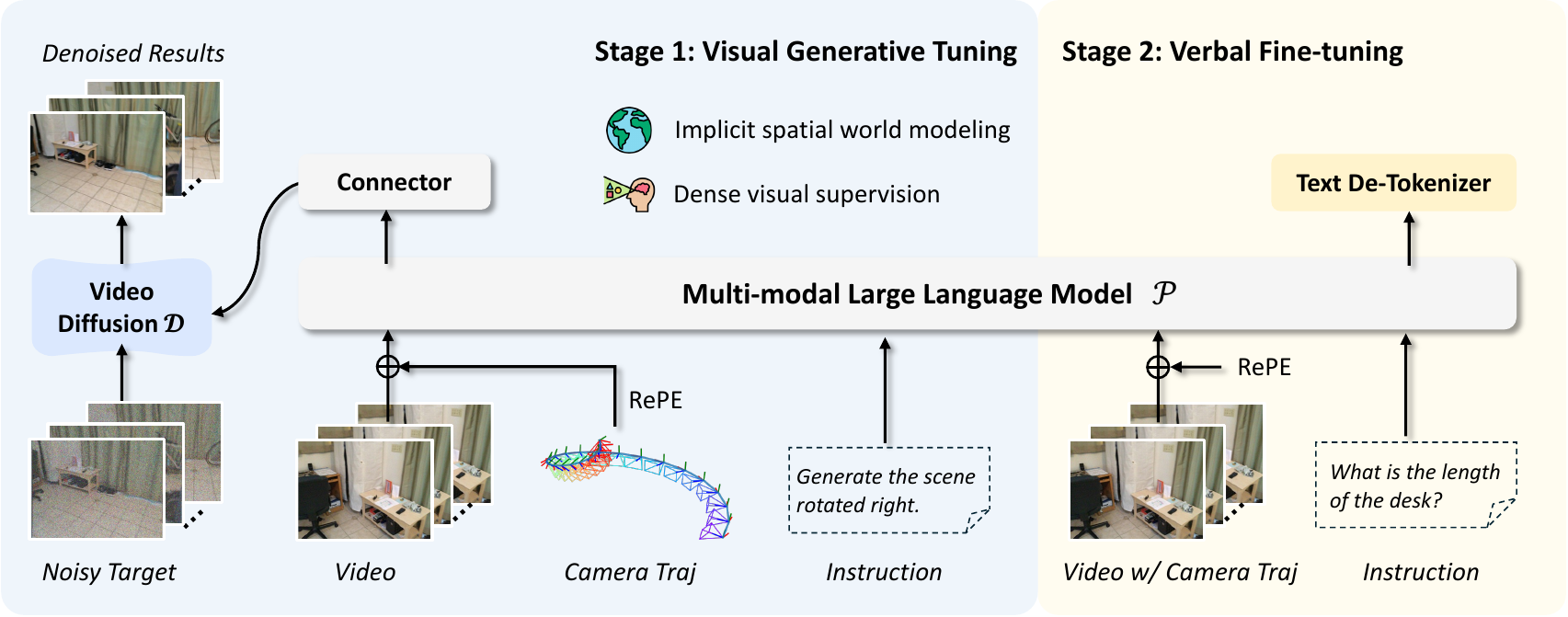}
     \caption{\textbf{An overview of MILO}, which consists of a MLLM $\mathcal{P}$ for multi-modal understanding, a video diffusion model $\mathcal{D}$ for visual feedback generation, and a connector in-between for dimensional adjustment. \textbf{The relative positional encoding (RePE)} takes the relative camera-pose variations as input and generate high-dimensional 3D-aware positional embeddings.}
	\label{fig:pipeline}
\end{figure*}
\section{Methodology} \label{sec:method}
\begin{figure}[t]
	\centering
        \includegraphics[width=0.49\textwidth]{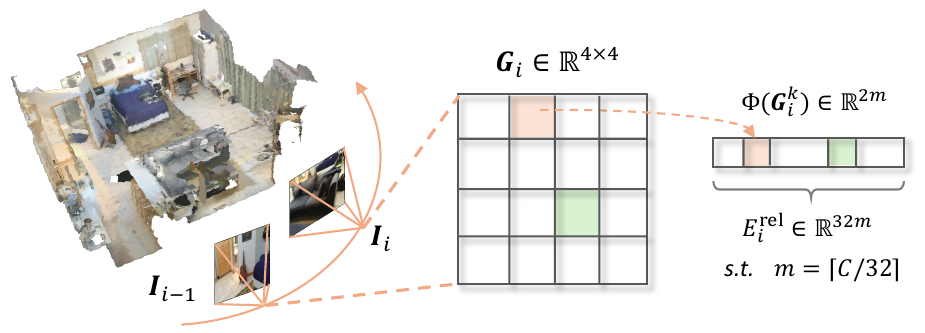}
     \caption{\textbf{Illustrations of RPE.} Give two consecutive frames $\boldsymbol{I}_{i}$ and $\boldsymbol{I}_{i-1}$, the relative geometric transformation matrix $\boldsymbol{G}_{i}$ is computed as Eq. \eqref{eq:1}. It is then converted into high-dimensional embeddings $\boldsymbol{E}_i^{\text{rel}}$ following Eq. \eqref{eq:3} and Eq. \eqref{eq:4}.}
	\label{fig:PE} 
\end{figure}
The overall architecture of MILO is illustrated in Figure \ref{fig:pipeline}. We first introduce the relative positional encoding in Section \ref{sec:3.1}. Then, in Section \ref{sec:3.2}, we describe the overall implicit spatial world modeling process. In Section \ref{sec:3.3}, we detail the construction process of the GeoGen dataset.

\subsection{Relative Positional Encoding} \label{sec:3.1}
Given a sequence of $N$ frames with their corresponding camera parameters $\{\left(\boldsymbol{I}_i, \boldsymbol{K}_i, \boldsymbol{T}_i^{c w}\right)\}_{i=1}^N$, where $\boldsymbol{K}_i \in \mathbb{R}^{3\times3}$ denotes the intrinsic matrix and $\boldsymbol{T}_i^{c w}=\left(\boldsymbol{R}_i^{c w}, \boldsymbol{t}_i^{c w}\right) \in S E(3)$ represents the camera pose in the world coordinate system, our goal is to model the relative camera transformations between adjacent frames rather than relying on an absolute coordinate frame. For each frame $i \ge 2$, the relative geometric transformation $\boldsymbol{G}_{i}$ between the current frame $\boldsymbol{I}_i$ and the previous one $\boldsymbol{I}_{i-1}$ is defined as follows:
\begin{equation}
\small
\boldsymbol{G}_{i} = \tilde{\boldsymbol{P}}_i \tilde{\boldsymbol{P}}_{i-1}^{-1} =
\begin{bmatrix}
\boldsymbol{K}_i & 0 \\
0 & 1
\end{bmatrix}
\boldsymbol{T}_i^{\text{cw}} (\boldsymbol{T}_{i-1}^{\text{cw}})^{-1}
\begin{bmatrix}
\boldsymbol{K}_{i-1}^{-1} & 0 \\
0 & 1
\end{bmatrix},
\label{eq:1}
\end{equation}
where $\tilde{\boldsymbol{P}}_i$ denotes the camera projection matrix in homogeneous form. For the first frame (\ie, $i = 1$), $\boldsymbol{G}_{1}$ is computed with respect to a reference camera pose by setting $\boldsymbol{K}_{\text{ref}} = I$ and $\boldsymbol{T}_{\text{ref}}^{\text{cw}} = I$:
\begin{equation}
\boldsymbol{G}_{1} = \tilde{\boldsymbol{P}}_1 \tilde{\boldsymbol{P}}_{\text{ref}}^{-1} =
\begin{bmatrix}
\boldsymbol{K}_{1} & 0 \\
0 & 1
\end{bmatrix}
\boldsymbol{T}_1^{\text{cw}}. 
\label{eq:2}
\end{equation}
Through this process, the relative geometric transformation $\boldsymbol{G}_{i} \in \mathbb{R}^{4\times4}$ encodes the relative rotation, translation, and intrinsic transformation between two adjacent frames. As shown in Figure \ref{fig:PE}, we project $\boldsymbol{G}_{i}$ into a high-dimensional representation following the sinusoidal positional encoding:
\begin{equation}
\small
\begin{aligned}
\Phi(\boldsymbol{G}_i^k)=
\Big[
&\sin\!\frac{\boldsymbol{G}_i^k}{\gamma^0},\, 
\cos\!\frac{\boldsymbol{G}_i^k}{\gamma^0},\, \ldots,\,\\[4pt]
&\sin\!\frac{\boldsymbol{G}_i^k}{\gamma^{2(m-1)/2m}},\,
\cos\!\frac{\boldsymbol{G}_i^k}{\gamma^{2(m-1)/2m}}
\Big]
\in \mathbb{R}^{2m},
\end{aligned}
\label{eq:3}
\end{equation}
\noindent where $\boldsymbol{G}_i^k$ denotes the $k$-th element of $\boldsymbol{G}_i$ and $\Phi(\boldsymbol{G}_i^k)$ is the corresponding sinusoidal positional encoding, $k \in [1,16]$. $\gamma$ is a frequency constant. Then, the final relative positional embedding $\boldsymbol{E}_i^{\text{rel}}$ for the $i$-th frame is obtained by concatenating  all the projected components:
\begin{equation}
\small
\boldsymbol{E}_i^{{\text{rel}}} = \text{concat}\big(\Phi(\boldsymbol{G}_i^1), \Phi(\boldsymbol{G}_i^2), \ldots, \Phi(\boldsymbol{G}_i^{16})\big) 
\in \mathbb{R}^{32m},
\label{eq:4}
\end{equation}
where we set $m=\lceil C/32\rceil$ to match the channel dimension of the semantic embeddings $\boldsymbol{E}_i^{\text{2D}}$ extract by CLIP \cite{radford2021learning}. Then the final 3D-aware representation $\boldsymbol{E}_i$ for the $i$-the frame is computed as follows:
\begin{equation}
\boldsymbol{E}_i^{\text{3D}} = \boldsymbol{E}_i^{\text{rel}} + \boldsymbol{E}_i^{\text{2D}}.
\end{equation}

\subsection{Implicit Spatial World Modeling} \label{sec:3.2}
The proposed implicit spatial world modeling is consisted of two consecutive training processes including the visual generative tuning under geometry-aware transformations and the verbal fine-tuning stage.

\begin{figure*}[t]
	\centering
        \includegraphics[width=0.99\textwidth]{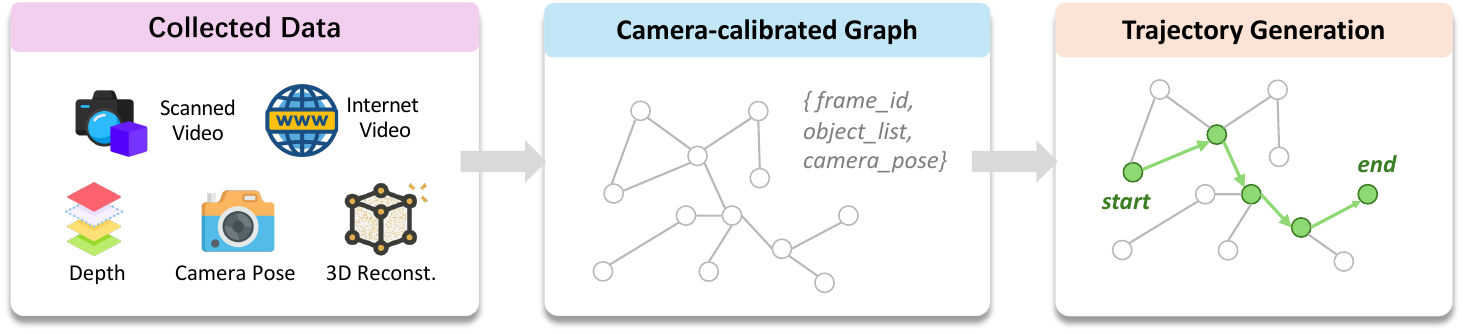}
     \caption{\textbf{Pipeline of trajectory generation.} Each video with geometric annotations is converted into a camera-calibrated graph based on the spatial connectivity between frames. The A* algorithm is employed to find the shortest path from the start point to the goal.}
	\label{fig:datasetGen}
\end{figure*}
\begin{figure*}[t]
    \centering
    \begin{subfigure}[b]{0.49\textwidth}
        \centering
        \includegraphics[width=0.97\textwidth]{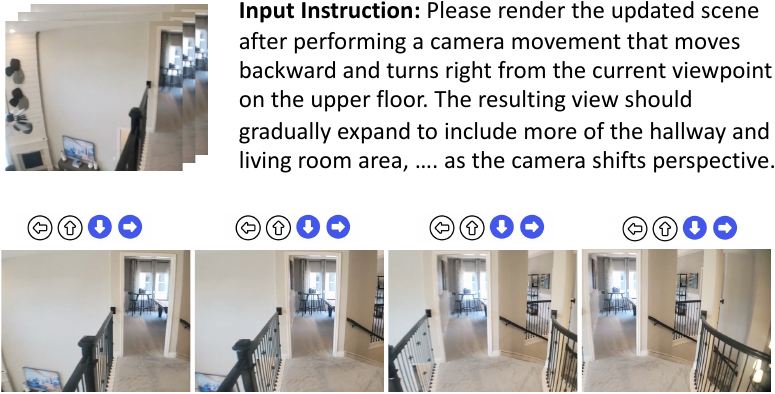}
        \caption{}
        \label{fig:datasetVis1}
    \end{subfigure}
    \begin{subfigure}[b]{0.49\textwidth}
        \centering
        \includegraphics[width=0.97\textwidth]{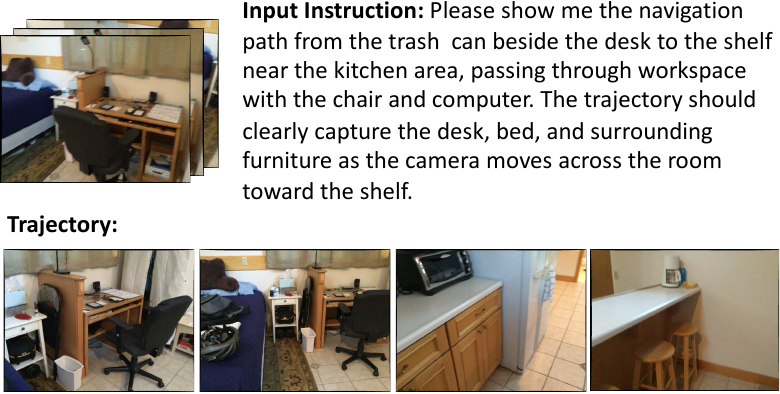}
        \caption{}
        \label{fig:datasetVis2}
    \end{subfigure}
    \caption{\textbf{Example cases} of (a) novel view synthesis and (b) trajectory generation in our GeoGen dataset.}
    \label{fig:datasetVis}
\end{figure*}

\paragraph{Visual Generative Tuning.} As illustrated in Figure~\ref{fig:teaser}, a video diffusion model $\mathcal{D}$ is used to reconstruct targets conditioned on the hidden visual embeddings of MLLM $\mathcal{P}$. Following the diffusion formulation, we adopt a denoising process that iteratively recovers clean latent tokens $z_0 = \mathcal{F}(\boldsymbol{I})$ from noisy tokens $z_t$, where $\boldsymbol{I}$ denotes the target visual sequences and $\mathcal{F}(\cdot)$ is implemented as a continuous VAE provided by FLUX \cite{labs2025flux1kontextflowmatching}. The diffusion model $\mathcal{D}$ is optimized to predict the noise $\epsilon$ given timestep $t$:
\begin{equation}
\mathcal{L}_{\text{MILO}} = 
\mathbb{E}_{t, \epsilon}\big[
\| \mathcal{D}(\boldsymbol{z}_t \mid \mathcal{P}(\boldsymbol{E}^{\text{3D}}, \boldsymbol{X^{\text{q}}}), t) - \epsilon \|_2^2
\big],
\label{eq:diffusion_loss}
\end{equation}
where $\boldsymbol{X^{\text{q}}}$ denotes the transformation instruction (\eg, \textit{``rotate right''}). $\epsilon \sim \mathcal{N}(0, 1)$ and $z_t$ is sampled from the standard forward diffusion process:
\begin{equation}
\boldsymbol{z}_t = \sqrt{\bar{\alpha}_t} \boldsymbol{z}_0 + \sqrt{1 - \bar{\alpha}_t}\,\epsilon, 
\quad 
\bar{\alpha}_t = \prod_{i=1}^{t} (1 - \beta_i),
\label{eq:diffusion_forward}
\end{equation}
with $\{\beta_t\}_{t=1}^{T}$ denoting the variance schedule. 

\paragraph{Verbal Fine-tuning.} After visual generative tuning, we conduct verbal fine-tuning with the auto-regressive loss.
\begin{equation}
\mathcal{L}_{\text{text}} =
\mathbb{E}_{\boldsymbol{X}^{\text{a}}_i}
\big[
-\log\,p_\theta
\big(
\boldsymbol{X}^{\text{a}}_i \mid 
\mathcal{P}
(\boldsymbol{E}^{\text{3D}}, \boldsymbol{X^{\text{q}}}),
\boldsymbol{X}^{\text{a}}_{<i}
\big)
\big],
\label{eq:text_loss}
\end{equation}
where $\boldsymbol{X^{\text{q}}}$ is the fine-tuning instruction (\eg, \textit{``what is the relative direction''}) and $\boldsymbol{X}^{\text{a}}_i$ is the generated responses.

\begin{table*}[t]
\centering
\renewcommand{\arraystretch}{1.1}
\setlength{\tabcolsep}{8pt} 
\caption{\textbf{Experimental results on 3D scene understanding benchmarks.} ``Expert Models" train a separate model for each task, whereas ``Generalist MLLMs" train a single model for all tasks. $^{\dagger}$ denotes reproduced results. Our MILO is built upon Video-3D LLM \cite{zheng2025video}.}
\resizebox{0.99\linewidth}{!}{
\begin{tabular}{lccccccccccc}
\toprule
\multirow{2}{*}{\textbf{Method}}
& \multirow{2}{*}{\makecell{\textbf{Point}\\\textbf{Encoder}}}
& \multicolumn{2}{c}{\textbf{ScanRefer}} 
& \multicolumn{2}{c}{\textbf{Multi3DRefer}} 
& \multicolumn{2}{c}{\textbf{Scan2Cap}}  
& \multicolumn{2}{c}{\textbf{ScanQA}} 
& \textbf{SQA3D} \\ 
\cmidrule(lr){3-4} \cmidrule(lr){5-6} \cmidrule(lr){7-8} \cmidrule(lr){9-10} \cmidrule(lr){11-11}
& & \textbf{Acc$@$0.25} & \textbf{Acc$@$0.5} & \textbf{F1$@$0.25} & \textbf{F1$@$0.5} & \textbf{C$@$0.5} & \textbf{B-4$@$0.5} & \textbf{C} & \textbf{EM} & \textbf{EM} \\
\midrule
\multicolumn{11}{l}{\textit{\textbf{Expert Models}}} \\
3D-VLP \cite{jin2023context}  & \cmark & 51.4  & 39.5 & --- & --- & 54.9 & 32.3 & 67.0 & 21.7 & ---\\
3D-VisTA \cite{zhu20233d}  & \cmark & 50.6 & 45.8 & --- & --- & 61.6 & 34.0 & 69.6 & 22.4 & 48.5 \\
3DJCG \cite{cai20223djcg} & \cmark &  49.6 & 37.3 & --- & 26.6 & 49.5 & 31.0 & --- & ---& ---\\
\hdashline
\multicolumn{11}{l}{\textit{\textbf{Generalist MLLMs}}} \\
3D-LLM \cite{hong20233d} & \cmark     & 30.3 & --   & --   & --   & --   & --   & 69.4 & 20.5 & 49.4   \\
Chat-3D v2 \cite{wang2023chat}     & \cmark             & 42.5 & 38.4 & 45.1 & 41.6 & 63.9 & 31.8 & 87.6 & --   & 54.7 \\
LL3DA \cite{chen2024ll3da}      & \cmark                & --   & --   & --   & --   & 62.9 & 36.0 & 76.8 & --   & --   \\
SceneLLM \cite{fu2024scene}    & \cmark               & --   & --   & --   & --   & --   & --   & 80.0 & 27.2 & 53.6 \\
LEO \cite{huang2023embodied}         & \cmark               & --   & --   & --   & --   & 72.4 & 38.2 & 101.4& 21.5 & 50.0 \\
Grounded 3D-LLM \cite{chen2024grounded}    & \cmark        & 47.9 & 44.1 & 45.2 & 40.6 & 70.6 & 35.5 & 72.7 & --   & --   \\ 
PQ3D \cite{zhu2024unifying}        & \cmark                & 57.0 & 51.2 & --   & 50.1 & 80.3 & 36.0 & --   & --   & 47.1 \\
ChatScene \cite{zhang2024chatscene}     & \cmark             & 55.5 & 50.2 & 57.1 & 52.4 & 77.1 & 36.3 & 87.7 & 21.6 & 54.6 \\
Ross3D$^{\dagger}$ \cite{wang2025ross3d} & --- & 60.6 & 54.1 & 58.8 & 53.8 & 82.2 &  46.4 & 105.1 & 30.6 & 60.6\\
LLaVA-3D \cite{zhu2024llava}      & ---             & 54.1 & 42.4 & --   & --   & 79.2 & 41.1 & 91.7 & 27.0 & 55.6 \\
Inst3D-LLM \cite{yu2025inst3d}    & \cmark    & 57.8 & 51.6 & 58.3 & 53.5 & 79.7 & 38.3 & 88.6 & 24.6 & --   \\
3D-LLaVA \cite{deng20253d}    & \cmark   & 51.2 & 40.6 & -- & -- & 78.8 & 36.9  & 92.6 & -- & 54.5 \\
Video-3D LLM \cite{zheng2025video}   & ---            & 58.1 & 51.7 & 58.0 & 52.7 & 83.8 & 41.3 & 102.1 & 30.1  & 58.6 \\
\textbf{MILO (Ours)}    & ---  & \sota{61.3} & \sota{54.7}  & \sota{59.4} & \sota{54.2} & \sota{85.6} & \sota{47.5} & \sota{107.3} & \sota{31.0} & \sota{60.9} \\
\bottomrule
\end{tabular}
}
\label{tab:resScanNet}
\end{table*}


\subsection{GeoGen Dataset} \label{sec:3.3}

To facilitate training, we curate GeoGen, a large-scale dataset that provides the outcome results under the geometric transformations. By statistics, our GeoGen dataset consists of 2,241 videos and 267,827 annotated observation-action-outcome triplets.

\noindent \textbf{Data Acquisition.} To ensure broad coverage, we collect video data from both scanned 3D assets and Internet videos. The former provides accurate 3D annotations for use, while the latter offers scalable and diverse in-the-wild scene data:
\begin{itemize}
\item \emph{Scanned 3D assets}: We integrate public datasets from ScanNet \cite{dai2017scannet} and ScanNet++ \cite{yeshwanth2023scannetplus}, which consist of RGB-D videos of indoor scenes annotated with 3D camera poses, reconstructed surface meshes, and instance-level semantic labels; 

\item \emph{Internet videos}: We leverage the RoomTour3D \cite{han2025roomtour3d} dataset since it comprises geometric annotations including camera trajectories, relative depth maps, object tags, and bounding boxes.
\end{itemize}

\noindent \textbf{Data Annotation.} To provide high-quality geometric-aware transformation instructions, we design the annotation process from two perspectives: 
\begin{itemize}
\item \emph{Novel view synthesis:} As shown in Figure \ref{fig:datasetVis1}, given a reference frame and directional instructions, the model outputs corresponding video frames. To avoid abrupt scene transitions, we first partition each video into separate shots using PySceneDetect \cite{PySceneDetect}. According to the camera pose information of the collected video data, we can easily derive relative camera positional relationships (\eg, forward/backward, left/right) between frames to construct the primary  instructions.

\item \emph{Trajectory generation:} As shown in Figure \ref{fig:datasetVis2}, given a complete input video, the model is required to predict the trajectory from a start point to an end point, necessitating a comprehensive understanding of the scene's geometric structure. To achieve this, we propose a \emph{camera-calibrated graph} based annotation method. Specifically, each video frame is treated as a node in the graph, and the edges between nodes are formed if their camera distance falls below a predefined threshold (\ie, 0.5m) and no obstructions are detected (verified via 3D point clouds and depth maps). By randomly selecting start and end nodes, we apply A$^{*}$ search \cite{hart1968formal} to determine the shortest path, with the sequence of traversed nodes constituting the trajectory. The primary transformation instruction is formulated as: ``Please show me the path from \{object in start frame\} to \{object in end frame\}." 
\end{itemize}
Finally, the primary instruction–answer pairs and corresponding videos are fed to GPT-4o \cite{gpt4o} to rephrase the instructions for greater diversity and richer scene details.


 

\section{Experiment} \label{sec:exp}

\subsection{Experiment Settings} 

\noindent \textbf{Baseline and Datasets.} Our proposed implicit spatial world modeling and relative positional encoding are agnostic to MLLMs. Therefore, we select three baseline models and conduct experiments upon them.
\begin{itemize}
\item \textbf{Video-3D LLM \cite{zheng2025video}:} This is one of the early attempts at 3D scene understanding. We replace the original 3D position encoding in Video-3D LLM with our proposed RePE and apply the visual generative post-training paradigm (\ie, Stage 1). Then, we follow the official settings of Video-3D LLM for fine-tuning (\ie, Stage 2). The experimental datasets include ScanRefer \cite{chen2020scanrefer} and Multi3DRefer \cite{zhang2023multi3drefer} for spatial referential understanding, Scan2Cap \cite{chen2021scan2cap} for 3D scene captioning, and ScanQA \cite{azuma2022scanqa} and SQA3D \cite{ma2022sqa3d} for spatial question answering.  

\item \textbf{VG-LLM \cite{zheng2025learning}:} It is a recent work that leverages VGGT \cite{wang2025vggt} to augment MLLMs' inputs. We firstly perform visual generative post-training (\ie, Stage 1). Following the original configuration, we use a sampled dataset from SPAR-7M \cite{zhang2025flatland} and LLaVA-Hound split of LLaVA-Video-178K \cite{zhang2024video} as the training dataset for the fine-tuning of stage 2. VSI-Bench \cite{yang2025thinking} is used for evaluation.

\item \textbf{RoboRefer \cite{zhou2025roborefer}}: This is designed for embodied spatial referring. Specifically, we firstly incorporate visual generative tuning into the depth alignment stage and then follow the official fine-tuning recipes. RefSpatial-Bench \cite{zhou2025roborefer} is used for evaluation. 
\end{itemize}

Notably, since VG-LLM \cite{zheng2025learning} and RoboRefer \cite{zhou2025roborefer} do not explicitly utilize 3D coordinates for input, we also omit the use of our RePE for a fair comparison.

\noindent \textbf{Implementation Details.} For visual generative tuning, the training step, batch size, and learning rate are set to 300, 16, and 1e-5 for the Video-3D LLM baseline \cite{zheng2025video}; 500, 256, and 5e-6 for the VG-LLM baseline \cite{zheng2025learning}; and 11534, 448, and 1e-3 for the RoboRefer baseline \cite{zhou2025roborefer}, respectively. During this stage, the MLLM backbone, the diffusion network, and the connector in-between are set to trainable while the visual encoder is frozen. All experiments are conducted on 8 NVIDIA A100 GPUs.  For the verbal fine-tuning stage, we follow the original training recipes of the baselines \cite{zheng2025video,zheng2025learning,zhou2025roborefer}.

\begin{table*}[t]
\belowrulesep=0pt
\aboverulesep=0pt
\renewcommand\arraystretch{1.1}
\setlength{\tabcolsep}{5pt} 
    \centering
    \caption{\textbf{Experimental results on VSI-Bench.} $^{\dagger}$ denotes the reproduced results. Our MILO is build upon VG-LLM \cite{zheng2025learning} with the same training dataset configurations.} 
    \resizebox{0.99\linewidth}{!}{
    \begin{tabular}{r|c|c|cccccccc}
    \toprule
    & & &
    \rotatebox{30}{\textbf{Obj. Count}} &
    \rotatebox{30}{\textbf{Abs. Dist.}} &
    \rotatebox{30}{\textbf{Obj. Size}} & 
    \rotatebox{30}{\textbf{Room Size}} &
    \rotatebox{30}{\textbf{Rel. Dist.}} &
    \rotatebox{30}{\textbf{Rel. Dir.}} &
    \rotatebox{30}{\textbf{Route Plan}} &
    \rotatebox{30}{\textbf{Appr. Order}} \\
    \makecell[c]{\textbf{Model}} & \textbf{\#Param} & \textbf{Avg.} & \multicolumn{4}{c}{\textbf{Numerical Answer}} & \multicolumn{4}{c}{\textbf{Multiple-Choice Answer}} \\
    \hline
    \rowcolor{lightgray!10}
    \multicolumn{1}{l|}{\textcolor{black}{\textit{General MLLMs}}} & & & & & & & & & & \\
    GPT-4o \cite{gpt4o} & - & 34.0 & 46.2 & 5.3 & 43.8 & 38.2 & 37.0 & 41.3 & 31.5 & 28.5 \\
    Gemini-1.5-Flash \cite{team2024gemini} & - & 42.1 & 49.8 & 30.8 & 53.5 & {54.4} & 37.7 & 41.0 & 31.5 & 37.8 \\
    Gemini-1.5-Pro \cite{team2024gemini} & - & 45.4 & {56.2} & {30.9} & {64.1} & 43.6 & {51.3} & {46.3} & {36.0} & 34.6 \\
    InternVL2-8B \cite{chen2024far} & 8B & 34.6 & 23.1 & {28.7} & 48.2 & {39.8} & 36.7 & 30.7 & 29.9 & 39.6 \\
    InternVL2-40B \cite{chen2024far} & 40B & 36.0 & 34.9 & 26.9 & 46.5 & 31.8 & 42.1 & 32.2 & 34.0 & 39.6 \\
    LLaVA-NeXT-7B \cite{li2024llavanext-strong}& 7B & 35.6 & 48.5 & 14.0 & 47.8 & 24.2 & {43.5} & 42.4 & 34.0 & 30.6 \\
    LLaVA-NeXT-72B \cite{li2024llavanext-strong}& 72B & 40.9 & {48.9} & 22.8 & 57.4 & 35.3 & 42.4 & 36.7 & {35.0} & \sota{48.6} \\
    LLaVA-OV-7B \cite{li2024llava}& 7B & 32.4 & 47.7 & 20.2 & 47.4 & 12.3 & 42.5 & 35.2 & 29.4 & 24.4 \\
    LLaVA-OV-72B \cite{li2024llava}& 72B & 40.2 & 43.5 & 23.9 & {57.6} & 37.5 & 42.5 & 39.9 & 32.5 & 44.6  \\
    Qwen2.5-VL-7B \cite{bai2025qwen2}& 7B & 33.0 & 40.9 & 14.8 & 43.4 & 10.7 & 38.6 & 38.5  & 33.0 & 29.8 \\
    Qwen2.5-VL-72B \cite{bai2025qwen2}& 72B & 37.0 & 25.1 & 29.3 & 54.5 & 38.8 & 38.2 & 37.0  & 34.0 & 28.9 \\
    \hline
    \rowcolor{lightgray!5}
    \multicolumn{1}{l|}{\textcolor{black}{\textit{Spatial MLLMs}}} & & & & & & & & & & \\
    SPAR \cite{zhang2025flatland}& 8B & 41.1 & - & - & - & - & - & - & -  & - \\

    Video-R1 \cite{feng2025video}& 7B & 37.1 & - & - & - & - & - & - & -  & - \\
    vsGRPO-V \cite{liao2025improved}& 7B & 40.7 & 59.9 & 29.6 & 50.8 & 48.3 & 35.4 & 35.6 & 34.0 & 31.5 \\
    SpaceR \cite{ouyang2025spacer}& 7B & 45.6 & - & - & - & - & - & - & -  & - \\
    VG-LLM$^{\dagger}$ \cite{zheng2025learning}& 8B & 59.5 & 69.9 & 51.4  & 66.5  &64.2 & 63.5  & 79.6  & 45.4 & 28.3 \\
    MILO (Ours) & 8B & \sota{61.7} & \sota{70.1} & \sota{55.1} & \sota{66.8} & \sota{65.9} & \sota{66.8} & \sota{85.3} & \sota{46.9} & 28.2 \\
    \bottomrule
    \end{tabular}
    }

\label{tab:resVSIBench}
\end{table*}


\begin{table*}[t]
\belowrulesep=0pt
\aboverulesep=0pt
\caption{\textbf{Experimental results on RefSpatial-Bench} including the splits of location, placement, and unseen compositional spatial relation. The \sota{red} and \underline{underlines} values represent the top-1 and top-2 accuracies, respectively. * denotes the reproduced results.
}
\centering
\renewcommand\arraystretch{1.1}
\setlength{\tabcolsep}{6pt}
\resizebox{\linewidth}{!}{
\begin{tabular}{l|ccccccccc}
\toprule
\multirow{3}{*}{RefSpatial-Bench}
& \cellcolor{roborefgray} \textit{Proprietary Models}
& \multicolumn{4}{c}{\cellcolor{roborefgreen}\textit{Referring Specialist Models}}   
& \multicolumn{2}{c}{\cellcolor{roborefblue}{RoboRefer} } 
& \multicolumn{1}{c}{\cellcolor{roborefred}{MILO} \textit{(Ours)}} \\

& \cellcolor{roborefgray}Gemini-2.5-Pro
& \cellcolor{roborefgreen}SpaceLLaVA
& \cellcolor{roborefgreen}RoboPoint
& \cellcolor{roborefgreen}Molmo-7B
& \cellcolor{roborefgreen}Molmo-72B
& \cellcolor{roborefblue}2B-SFT
& \cellcolor{roborefblue}2B-SFT *
& \cellcolor{roborefred}2B-SFT \\



\midrule

Location  & 46.96 & 5.82 & 22.87 & 21.91 & 45.77 & \underline{47.00} & \underline{47.00} & \sota{48.00}\\
Placement  & 24.21 & 4.31 & 9.27 & 12.85 & 14.74 & \sota{48.00} & \underline{46.00} & \underline{46.00} \\
Unseen  & 27.14 & 4.02 & 8.40 & 12.23 & 21.24 & 33.77 & \underline{35.06} & \sota{36.36} \\

\bottomrule[1pt]
\end{tabular}}
\label{tab:resRefSpatial}
\end{table*}


\subsection{Experimental Comparisons} 

\noindent \textbf{Results on 3D scene understanding benchmarks.} We compare our MILO against both specialized expert models and generalist MLLMs on five challenging benchmarks. As summarized in Table \ref{tab:resScanNet}, MILO achieves superior performance across all tasks while requiring no explicit 3D point cloud encoder. Notably, compared to our direct baseline Video-3D LLM \cite{zheng2025video}, MILO shows consistent improvements across all metrics (e.g., +3.2\% Acc@0.25 on ScanRefer \cite{chen2020scanrefer}, +6.2\% B-4@0.5 on Scan2Cap \cite{chen2021scan2cap}), validating the effectiveness of our implicit spatial world modeling. 
\begin{table*}[t]
\centering
\renewcommand{\arraystretch}{1.1}
\setlength{\tabcolsep}{10pt} 
\caption{\textbf{Ablation studies. Exp \#2:} ``\emph{w/o} visual gen" denotes removing the visual generative tuning stage (\cf Section \ref{sec:3.2}). \textbf{Exp \#3 \& \#4:} ``\emph{novel view only}" indicates using only the novel view synthesis split from the GeoGen dataset, while ``\emph{trajectory only}" refers to using only the trajectory generation split (\cf Section \ref{sec:3.3}). \textbf{Exp \#5:} ``\emph{w/o} RePE" denotes using the vanilla absolute 3D coordinate–based positional encoding instead of our proposed relative positional encoding (\cf Section \ref{sec:3.1}).}
\resizebox{\linewidth}{!}{
\begin{tabular}{llccccccccccc}
\toprule
\multirow{2}{*}{\textbf{Exp}} & \multirow{2}{*}{\textbf{Method}}
& \multicolumn{2}{c}{\textbf{ScanRefer}} 
& \multicolumn{2}{c}{\textbf{Multi3DRefer}} 
& \multicolumn{2}{c}{\textbf{Scan2Cap}}  
& \multicolumn{2}{c}{\textbf{ScanQA}} 
& \textbf{SQA3D} \\ 
& & \textbf{Acc$@$0.25} & \textbf{Acc$@$0.5} & \textbf{F1$@$0.25} & \textbf{F1$@$0.5} & \textbf{C$@$0.5} & \textbf{B-4$@$0.5} & \textbf{C} & \textbf{EM} & \textbf{EM} \\
\midrule
\rowcolor{roborefred}
\#1 & MILO    & \sota{61.3}  & \sota{54.7} & \sota{59.4} & \sota{54.2} & \sota{85.6} & \sota{47.5} & \sota{107.3} & \sota{31.0} & \sota{60.9} \\
\#2 &\emph{w/o} visual gen & 59.6  & 52.9 & 58.2 & 52.9 & 85.1 & 46.6 & 99.4 & 28.2 & 60.2 \\
\#3 &novel view only  & 59.6  & 53.0 & 57.9 & 52.8 & 85.4 & 42.4 & 104.3 & 29.9 & 59.3 \\
\#4 &trajectory only & 59.4  & 52.9 & 57.8 & 52.6 & 82.5 & 46.0 & 103.1 & 29.3 & 60.1 \\
\#5 &\emph{w/o} RePE & 59.8  & 53.2 & 58.3 & 53.1 & 85.6 & 46.4 & 103.6 & 30.0 & 59.9\\
\bottomrule
\end{tabular}
}
\label{tab:ablationComponent}
\end{table*}

\noindent \textbf{Results on spatial reasoning benchmarks.} As shown in Table \ref{tab:resVSIBench}, our MILO achieves superior performance with an average accuracy of 61.7\%, outperforming the baseline VG-LLM \cite{zheng2025learning} by +2.2\% absolute improvement. In addition, MILO yields consistent improvements across almost all sub-tasks, with the largest gains observed in the sub-tasks of relative direction (+5.7\%), relative distance (+3.3\%), and absolute distance (+3.7\%). These results suggest that the proposed visual generative tuning yields more pronounced improvements on sub-tasks involving spatial direction and distance reasoning.

\noindent \textbf{Results on embodied referring benchmarks.} We further evaluate our MILO on the challenging RefSpatial-Bench, which contains three distinct splits of location, placement, and unseen compositional spatial relation. As reported in Table \ref{tab:resRefSpatial}, MILO achieves highly competitive performance across all three splits. Compared to the RoboRefer \cite{zhou2025roborefer} baseline, our MILO  achieves absolute gains of +1.00\% on the location split (48.00\% vs. 47.00\%) and +1.30\% on the challenging unseen split.

\subsection{Ablation Studies} 
\begin{table}[t]
\centering
\renewcommand{\arraystretch}{1.1}
\setlength{\tabcolsep}{10pt} 
\caption{\textbf{Evaluation of Grounded Attention Score (GAS) on RefSpatial-Bench.} GAS is defined as the attention mass allocated to the ground-truth visual region by the generated last response token. ``Acc" denotes the average accuracy (\%) of all the three splits of RefSpatial-Bench.}
\resizebox{0.9\linewidth}{!}{
\begin{tabular}{lcc}
\toprule
\textbf{Method} & \textbf{GAS} & \textbf{Acc} \\
\midrule
\emph{w/o} visual generative tuning  & 0.014 &   43.3 \\
\emph{w/} visual generative tuning   & \textbf{0.077}  & \textbf{44.0} \\
\bottomrule
\end{tabular}
}
\label{tab:gasRes}
\end{table}


\begin{figure*}[t]
	\centering
    \includegraphics[width=\textwidth]{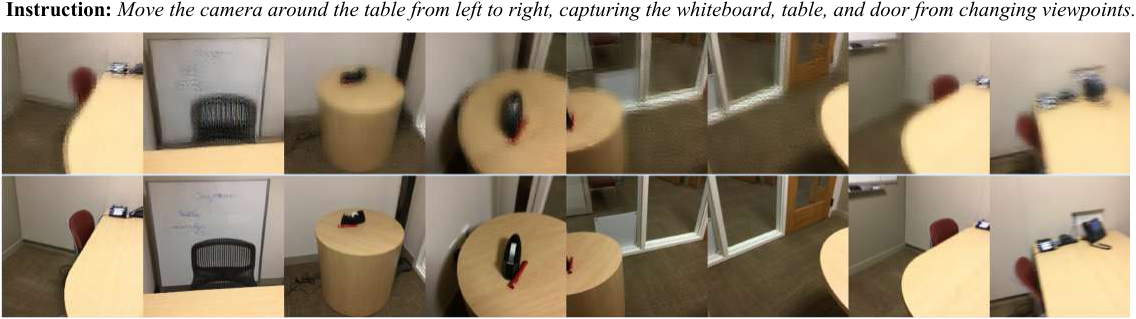}
     \caption{\textbf{Visual comparison} between generated results (top) and ground truth (bottom).}
	\label{fig:visGen}
\end{figure*}
\noindent \textbf{Motivation clarification.} Our approach is motivated by the visual illiteracy issue in current MLLMs, which often fail to accurately attend to relevant regions when generating responses. To systematically evaluate this, we design a \emph{grounded attention assessment} on RefSpatial-Bench \cite{zhou2025roborefer}. Specifically, we compute the attention distribution of the last response token over all visual tokens and calculate the sum of attention scores within the ground-truth mask region, termed the \emph{grounded attention score}. As shown in Table \ref{tab:gasRes}, the proposed visual generative tuning significantly improves both the grounded attention score and the overall accuracy, demonstrating its effectiveness in enhancing the spatial awareness of MLLMs during response generation. 

\noindent \textbf{Ablations on visual generative tuning.} We conduct ablation studies to validate the efficacy of our proposed visual generative tuning (\cf Exp \#1 \vs Exp \#2 in Table \ref{tab:ablationComponent}). The results consistently indicate that removing the visual generative tuning degrades performance, with a 1.7\% drop in Acc@0.25 on ScanRefer \cite{chen2020scanrefer} as a representative example.

\noindent \textbf{Ablations on GeoGen components.} As detailed in Section \ref{sec:3.3}, the curated GeoGen dataset comprises two distinct data types: novel view synthesis and trajectory generation. We conduct an ablation study to evaluate the contribution of each data type. Results of Exp \#3 and \#4 in Table \ref{tab:ablationComponent} demonstrate that both types of data contribute significantly to the final model performance.

\noindent \textbf{Ablations on RePE.}  We ablate the proposed RePE by replacing it with vanilla absolute 3D coordinate-based positional encoding \cite{zheng2025video}. The comparative results between Exp \#1 and Exp \#5 in Table \ref{tab:ablationComponent} clearly demonstrate the effectiveness of our proposed RePE, revealing a consistent performance improvement across all benchmarks.

\noindent \textbf{Visualizations.} Figure \ref{fig:visGen} presents examples of the visual generation results in the visual generative tuning stage, where MLLMs are required to generate novel views under camera movement instructions. As shown in Figure \ref{fig:visGen}, the generated views are generally consistent with the ground truth frames. Although some pixel-level artifacts remain, this is an expected outcome, as our visual generative tuning primarily focuses on establishing \emph{high-level} spatial world modeling and implicit spatial awareness for MLLMs, rather than pursuing \emph{pixel-level} photorealism. 

\section{Conclusions and Future Works} \label{sec:con}
In this work, we presented MILO, an implicit spatial world modeling paradigm that endows MLLMs with human-like spatial imagination and geometric awareness. Unlike conventional verbal descriptive tuning, MILO incorporates visual generative tuning to provide geometry-aware visual feedback, implicitly grounding symbolic reasoning in perceptual experience. To further enhance geometric understanding, we introduced RePE, a relative positional encoding scheme that captures inter-view camera-pose relationships in a coordinate-agnostic manner. Supported by the curated GeoGen dataset, MILO consistently improves spatial reasoning across multiple baselines and benchmarks. In future work, we plan to extend MILO toward real-world embodied settings, enabling on-policy interaction with physical environments through active visual imagination.

\section{Appendix}
This appendix provides additional experimental analyses and qualitative visualizations to complement the main paper. 

Specifically, the experiment results include:
\begin{itemize}[leftmargin=13pt, itemsep=3pt]
    \item Human evaluations of visual illiteracy issue.
    \item Statistics of GeoGen dataset.
\end{itemize}
\vspace{1mm}
The visualization results include:
\begin{itemize}[leftmargin=13pt, itemsep=3pt]
    \item Visualization of grounded attention assessment. 
    \item Visualization of GeoGen dataset.
    \item Visualization of generation results.
\end{itemize}

\subsection{More Experimental Results}

\noindent \textbf{Human evaluations of visual illiteracy issue.} To validate the visual illiteracy in current MLLMs, we have conducted a grounded attention assessment on RefSpatial-Bench \cite{zhou2025roborefer} in Section 4.3 of the main paper. Since the ScanNet series datasets \cite{chen2020scanrefer,zhang2023multi3drefer,azuma2022scanqa,chen2021scan2cap} do not provide such affordance annotations, we design a human evaluation on the ScanQA dataset \cite{azuma2022scanqa} to examine whether MLLMs focus on relevant visual regions when generating answers. Specifically, we randomly sample 200 instances from ScanQA and compute the attention distribution of the final response token across all visual tokens. Participants are required to judge whether the attention scores are concentrated on regions related to the question, with evaluation options including: focused, partially focused, and not focused.

\begin{figure}[t]
	\centering
        \includegraphics[width=0.49\textwidth]{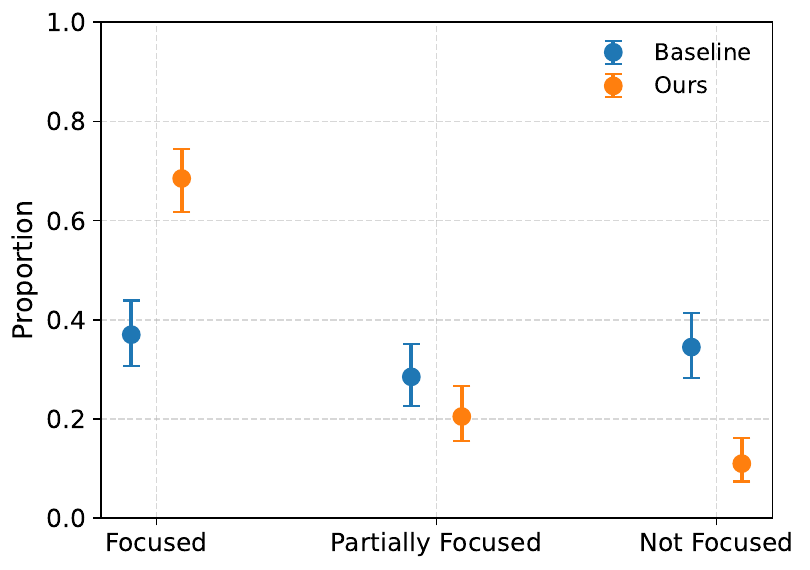}
     \caption{\textbf{Human evaluation of visual attention distribution} between Video-3D LLM baseline \cite{zheng2025video} and our MILO. Participants are required to judge whether the visual attention distributions are focused on the question-relevant regions.}
	\label{fig:humanEval}
\end{figure}

As illustrated in Figure \ref{fig:humanEval}, our proposed MILO demonstrates a significantly higher proportion of focused attention compared to the baseline Video-3D LLM \cite{zheng2025video}. This indicates that our proposed implicit spatial world modeling effectively mitigates visual illiteracy by directing the model's attention to question-relevant regions.

\begin{table}[t]
\centering
\renewcommand{\arraystretch}{1.1}
\caption{\textbf{Key statistics} of GeoGen dataset.}
\begin{tabular}{lr}
        \toprule
        \textbf{Statistics of GeoGen} & \textbf{Value} \\
        \midrule
        Number of videos & 2,241 \\
        \quad $-$ Scanned 3D assets & {1513} \\
        \quad $-$ Internet videos & {728} \\
        Video Length (Seconds, avg/max) & {123.1} / {332.3} \\
        Total annotations triplets & 267,827 \\
         \quad $-$ Scanned 3D assets & {256187} \\
        \quad $-$ Internet videos & {11640} \\
        Instruction Length (avg/max) & {32.4} / {240} \\
        \bottomrule
\end{tabular}
\label{tab:statistics}
\end{table}

\begin{figure*}[t]
	\centering
        \includegraphics[width=\textwidth]{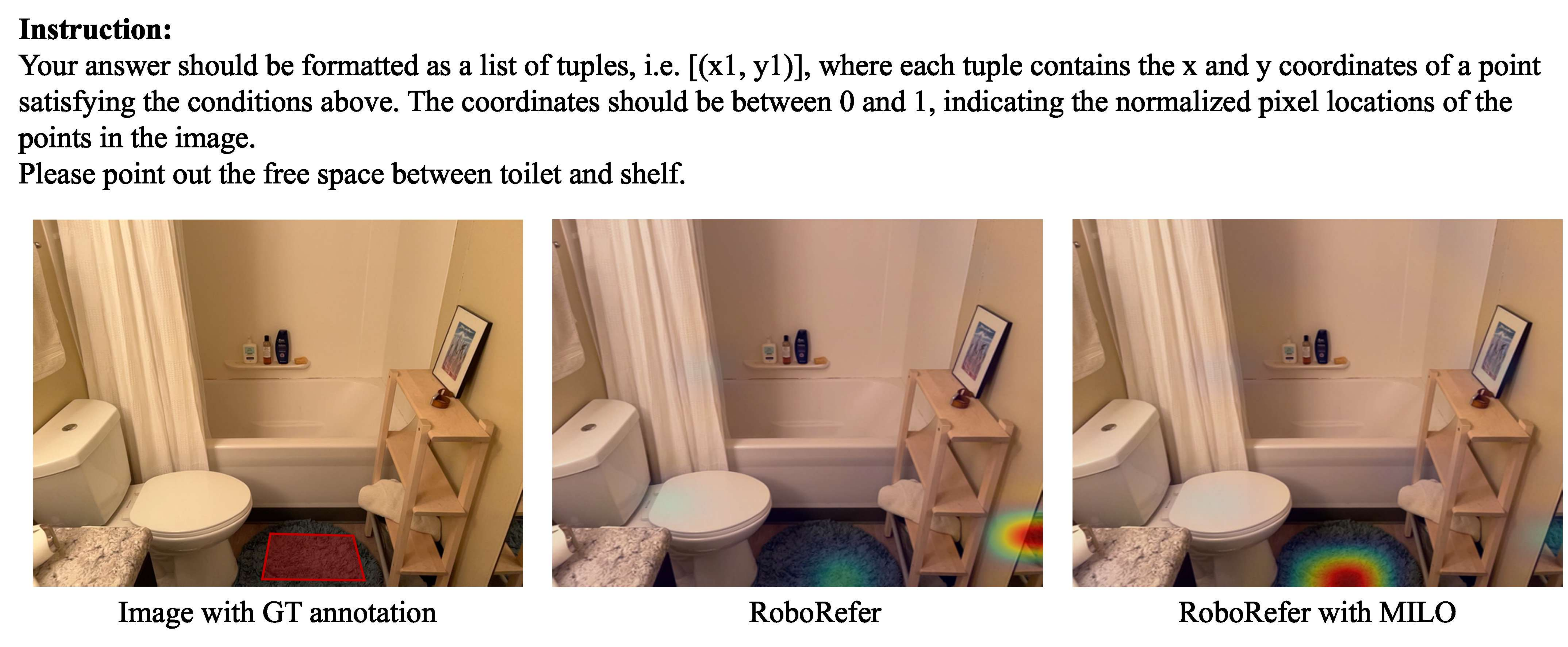}
        \includegraphics[width=1\textwidth]{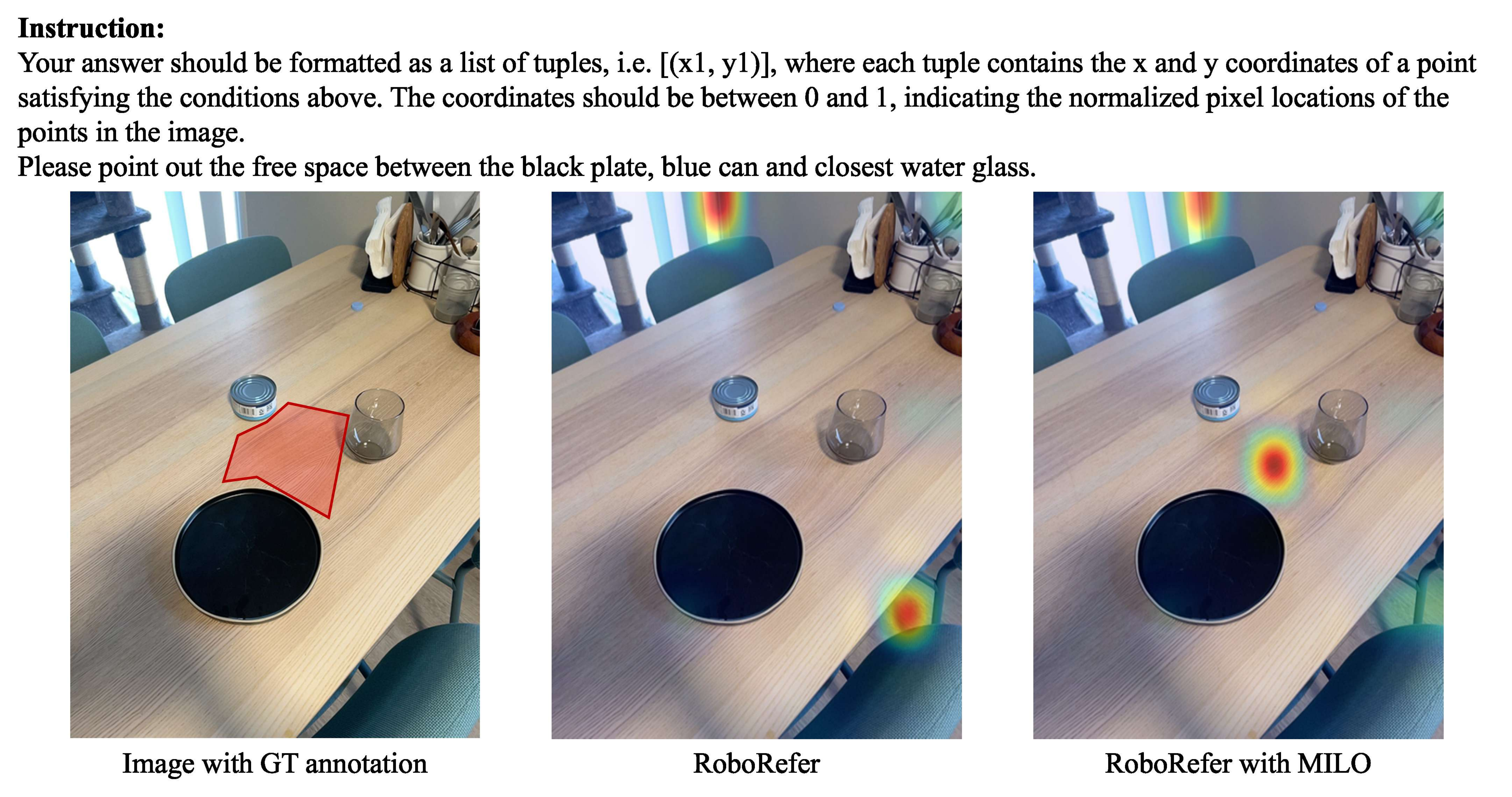}
     \caption{\textbf{Qualitative examples of grounded attention assessment} on RefSpatial-Bench \cite{zhou2025roborefer}, illustrating the attention distribution of the last response token over all the visual tokens.}
     \label{fig:groundedAtten}
\end{figure*}

%


\noindent \textbf{Statistics of GeoGen dataset.} Table \ref{tab:statistics} summarizes the key statistics of the GeoGen dataset, which contains 2,241 videos including 1,513 scanned 3D asset scenes and 728 Internet videos. The average video duration is 123.1 s, with a maximum of 332.3 s, providing sufficient temporal context for spatial reasoning. In total, GeoGen includes 267,827 observation–action–outcome triplets. These statistics demonstrate that GeoGen offers large-scale and diverse annotations for the proposed visual generative tuning.


\subsection{More Visualization}

\noindent \textbf{Visualization of grounded attention assessment.} In Table 5 of the main paper, we have presented the quantitative results of the grounded attention assessment on RefSpatial-Bench \cite{zhou2025roborefer}. Here, we qualitatively visualize several examples by plotting the attention distribution of the last response token over all visual tokens. As shown in Figure \ref{fig:groundedAtten}, the baseline model RoboRefer \cite{zhou2025roborefer} often fails to concentrate on the target regions specified in the instruction, leading to diffuse or misplaced attention. In contrast, our MILO demonstrates more focused and semantically aligned attention patterns that accurately correspond to the instructed spatial regions (\eg, the free space between objects). These visualizations provide intuitive evidence that MILO enhances the model's grounded visual reasoning and alleviates the issue of visual illiteracy.

\begin{figure*}[t]
	\centering
        \includegraphics[width=0.96\textwidth]{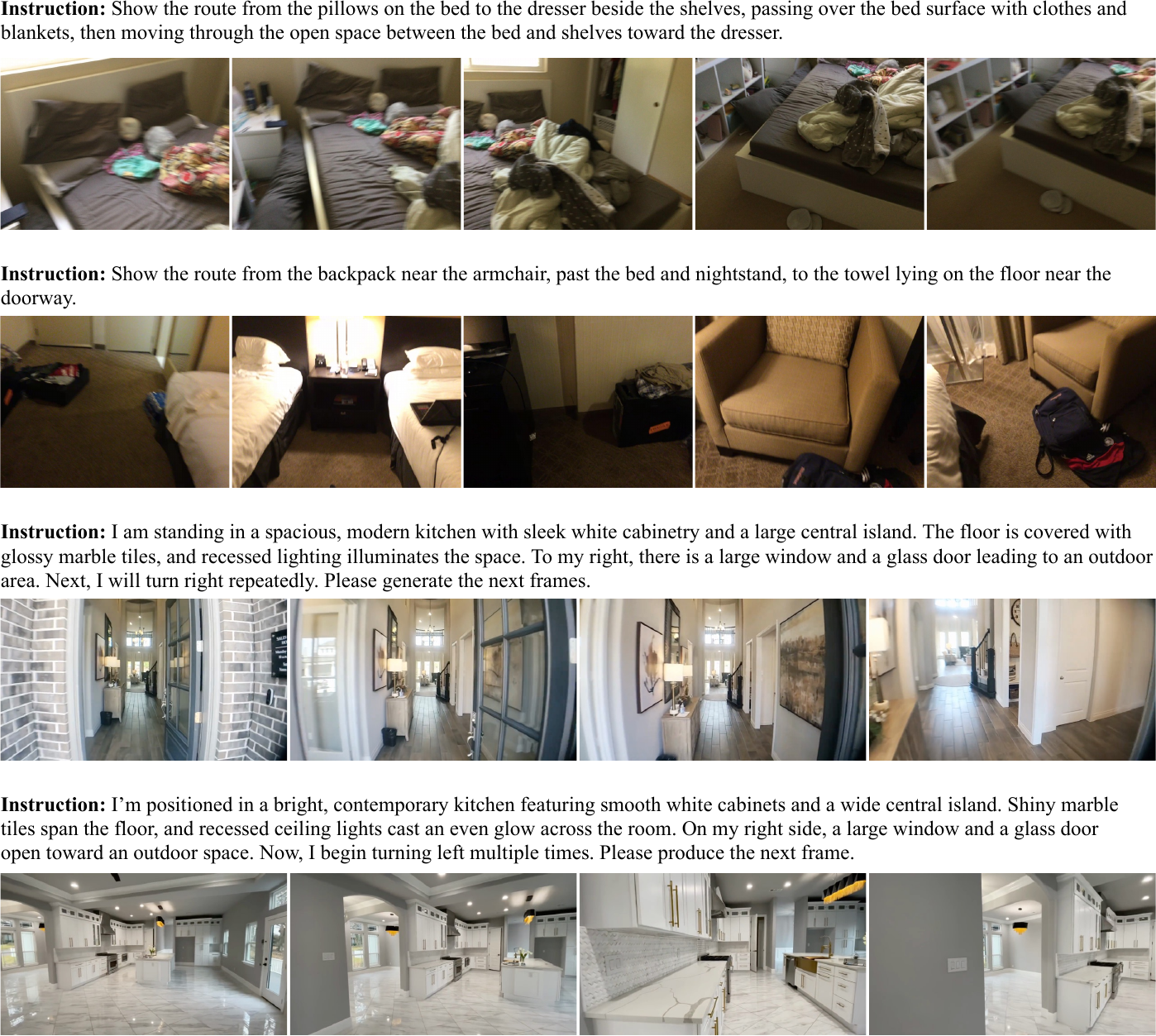}
     \caption{\textbf{Example cases of GeoGen dataset} including trajectory generation (top two cases) and novel view synthesis (bottom two cases).}
	\label{fig:datasetGenAppendix}
\end{figure*}

\noindent \textbf{Visualization of GeoGen dataset.} Figure \ref{fig:datasetGenAppendix} showcases representative examples from the GeoGen dataset, covering two core tasks: trajectory generation (top two cases) and novel view synthesis (bottom two cases). In the trajectory generation examples, the model follows natural-language instructions to produce coherent egocentric motion sequences that traverse indoor environments with fine-grained spatial details (\eg, ``from the pillows on the bed to the dresser"). In contrast, the novel view synthesis cases depict the model's ability to generate realistic frames conditioned on descriptive scene instructions (\eg, ``turn right repeatedly") within the kitchen layouts. Together, these examples highlight GeoGen's capacity to bridge linguistic spatial understanding and visual scene generation.

\begin{figure*}[t]
	\centering
        \includegraphics[width=1\textwidth]{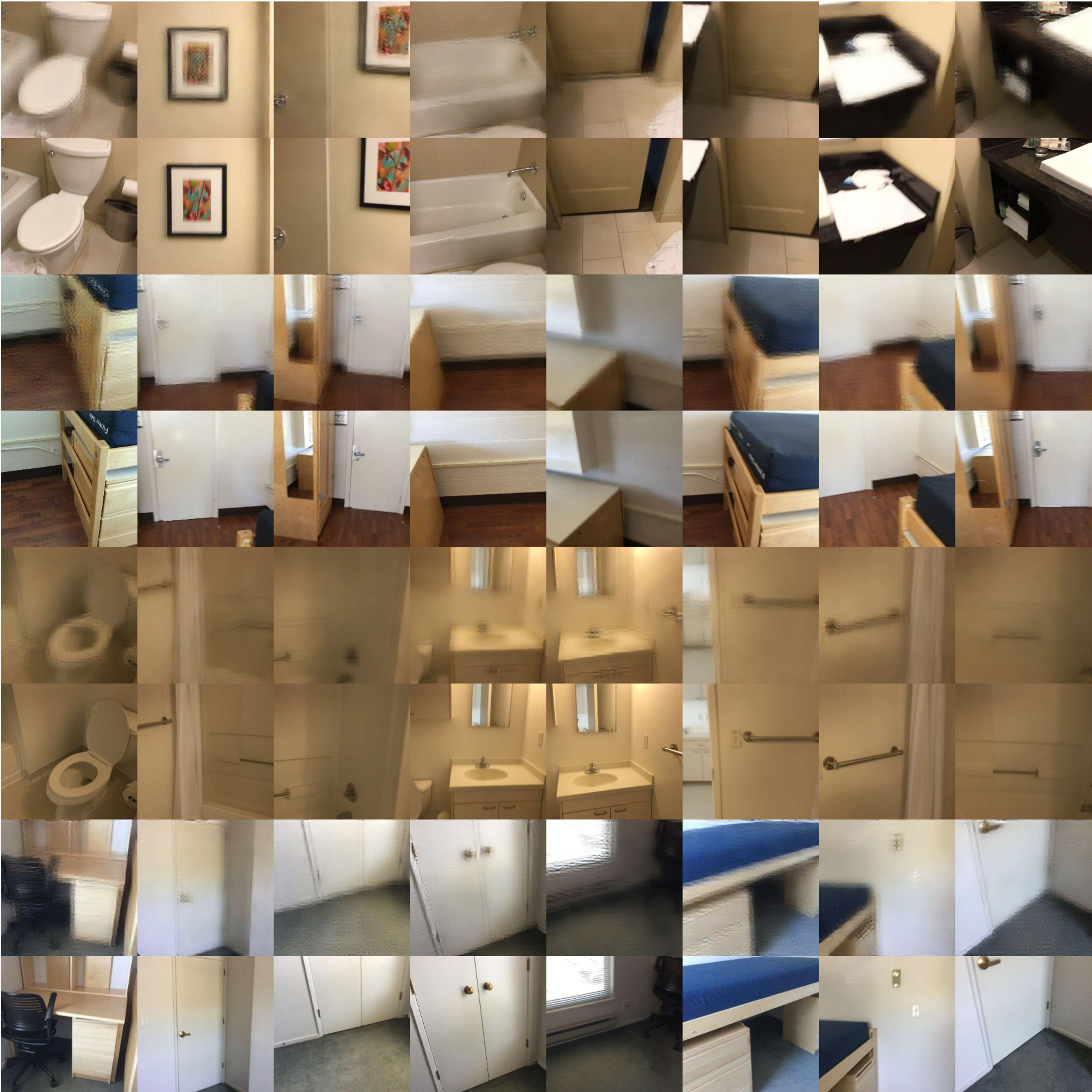}
     \caption{\textbf{Comparison between generated views (top) and ground-truth views (bottom)} across diverse indoor scenes.}
     \label{fig:visGenAppendix}
\end{figure*}

\noindent \textbf{Visualization of generation results.} Figure \ref{fig:visGenAppendix} presents qualitative examples generated by our model across diverse indoor scenes, including bathrooms, bedrooms, and offices. These results demonstrate that our approach produces geometrically consistent and semantically coherent views with plausible spatial layouts and textures.

{
    \small
    \bibliographystyle{ieeenat_fullname}
    \bibliography{references}
}


\end{document}